\documentclass[11pt]{article}
\usepackage{coling2020}
\usepackage{hyperref}
\usepackage{times}
\usepackage{url}
\usepackage{latexsym}

\usepackage{makecell}
\usepackage{ragged2e}
\usepackage[normalem]{ulem}
\usepackage{color,soul,bm}
\usepackage{graphicx}
\usepackage[inline]{enumitem}
\usepackage{amsmath}
\usepackage{amsfonts}
\usepackage{cleveref}
\usepackage{listings}
\usepackage{xcolor}
\usepackage{float}
\usepackage{booktabs}
\usepackage{multirow}
\usepackage{wrapfig}
\usepackage[final]{trackchanges}
\usepackage[font=normalsize,labelfont=bf]{caption}
 
\definecolor{codegreen}{rgb}{0,0.6,0}
\definecolor{codegray}{rgb}{0.5,0.5,0.5}
\definecolor{codepurple}{rgb}{0.58,0,0.82}
\definecolor{backcolour}{rgb}{0.95,0.95,0.92}

\definecolor{blue2}{rgb}{0.82,0.90,0.99}
\definecolor{blue3}{rgb}{0.40,0.75,0.98}
\definecolor{uva}{rgb}{0.90, 0.45, 0.00}
 
\lstdefinestyle{mystyle}{
    keywordstyle=\scshape\color{uva}
}

\lstnewenvironment{sqllstlisting}[1][]
 {%
  \lstset{ aboveskip=-1.9ex, belowskip=-2.7ex, #1}%
 }{}

\lstnewenvironment{gatelstlisting}[1][]
 {%
  \lstset{ basicstyle=\small, breaklines=false, aboveskip=-1.3ex, belowskip=-1.8ex, #1}%
 }{}
 
\usepackage{tikz}

\crefformat{section}{\S#2#1#3}
\crefformat{subsection}{\S#2#1#3}
\crefformat{subsubsection}{\S#2#1#3}
\crefrangeformat{section}{\S\S#3#1#4 to~#5#2#6}
\crefmultiformat{section}{\S\S#2#1#3}{ and~#2#1#3}{, #2#1#3}{ and~#2#1#3}

\newcommand{\flatgcv}[2]{\ensuremath{\rho_{\operatorname{l}}=#1, \rho_{\operatorname{c}}=#2}}
\newcommand{\txtunderset}[2]{$\underset{\text{#1}}{\text{#2}}$}

\usepackage{microtype}

\addeditor{YJ}

\colingfinalcopy 

\title{A Tale of Two Linkings: Dynamically Gating between Schema Linking and Structural Linking for Text-to-SQL Parsing}

\author{Sanxing Chen$^\diamondsuit$\ \ \ \ \ \ Aidan San$^\diamondsuit$\ \ \ \ \ \ Xiaodong Liu$^\clubsuit$\ \ \ \ \ \ Yangfeng Ji$^\diamondsuit$\\
  $^\diamondsuit$University of Virginia \\
  $^\clubsuit$Microsoft Research \\
  \texttt{\{sc3hn, aws9xm, yangfeng\}@virginia.edu} \\
  \texttt{xiaodl@microsoft.com} \\}

\date{}

\begin{document}
\maketitle
\begin{abstract}
In Text-to-SQL semantic parsing, selecting the correct entities (tables and columns) for the generated SQL query is both crucial and challenging; the parser is required to connect the natural language (NL) question and the SQL query to the structured knowledge in the database. We formulate two linking processes to address this challenge: \textit{schema linking} which links explicit NL mentions to the database and \textit{structural linking} which links the entities in the output SQL with their structural relationships in the database schema. Intuitively, the effectiveness of these two linking processes changes based on the entity being generated, thus we propose to dynamically choose between them using a gating mechanism. Integrating the proposed method with two graph neural network-based semantic parsers together with BERT representations demonstrates substantial gains in parsing accuracy on the challenging Spider dataset. Analyses show that our proposed method helps to enhance the structure of the model output when generating complicated SQL queries and offers more explainable predictions.
\end{abstract}

\section{Introduction}
\blfootnote{
    \hspace{-0.65cm}  
    This work is licensed under a Creative Commons 
    Attribution 4.0 International License.
    License details:
    \url{http://creativecommons.org/licenses/by/4.0/}.
}
Semantic parsing, which aims at mapping natural language (NL) utterances to computer understandable logic forms or programming languages, has been an active research topic in the field of natural language processing (NLP) for decades~\cite{zettlemoyer-2005-learning,liang-etal-2011-learning}. Although a variety of logic forms have been studied by researchers, Text-to-SQL has particularly attracted a large amount of attention due to the desire of natural language interfaces to database (NLIDB)~\cite{warren-pereira-1982-efficient,Zelle-1996-Learning,dong_learning_2019} for both scientific and industrial reasons. Recently, there is a growing interest in neural based Text-to-SQL semantic parsing, thanks to the development of new evaluation paradigms and datasets~\cite{iyer-etal-2017-learning,zhongSeq2SQL2017,yu-etal-2018-spider,yu-etal-2019-sparc}.

Text-to-SQL parsing requires strict \textit{structured prediction} due to its application scenario where the output SQL will be sent to an executor program directly. To enhance the capacity of an auto-regressive model to capture structural information, current state-of-the-art semantic parsers usually adopt a grammar-based decoder~\cite{xiao-etal-2016-sequence,yin-neubig-2017-syntactic,krishnamurthy-etal-2017-neural}. Rather than directly generating the tokens in a traditional sequence-to-sequence manner, grammar-based decoders produce a sequence of production rules to construct an abstract syntax tree (AST) of the corresponding SQL.
As the grammar constraints narrows down the search space to only grammatically valid ASTs,
those parsers can usually generate well-formed SQL skeletons~\cite{guo-etal-2019-towards,bogin-etal-2019-representing}.

\begin{table*}[t]
  \centering
  \begin{tabular}{p{0.02\linewidth}p{0.88\linewidth}}
    \toprule
    \multirow{3}{*}{1} 
                       & Q: \textit{For each \uline{continent}, list its \uline{id}, \uline{name}, and how many \uline{countries} it has?} \\
                       & \begin{sqllstlisting}
select t1.contid, t1.continent, count(*) from continents as t1 join countries as t2 on t1.contid = t2.continent group by t1.contid;
                       \end{sqllstlisting} \\ \midrule
    \multirow{3}{*}{2} & Q1: \textit{What is the average, minimum, and maximum \uline{age} of all \uline{singers} from \uwave{France}?}\\
                       & Q2: \textit{What is the average, minimum, and maximum \uline{age} for all \uwave{French} \uline{singers}?}\\
                       & \begin{sqllstlisting}
select avg(age), min(age), max(age) from singer where country = 'France';
                       \end{sqllstlisting} \\ \midrule
    \multirow{3}{*}{3} 
                       & Q: \textit{What is the \uline{first name} and \uline{gender} of the all the \uline{students} who have more than one \uline{pet}?}\\
                       & \begin{sqllstlisting}
select t1.fname, t1.sex from student as t1 join has_pet as t2 on t1.stuid = t2.stuid group by t1.stuid having count(*) > 1
                       \end{sqllstlisting} \\ \bottomrule
  \end{tabular}
  \caption{Several examples that are taken from the Spider dataset. Entity mentions are \uline{underlined} in NL questions. Q1 and Q2 are paraphrases of each other which should lead to the same SQL result. \uwave{Wavy underline} indicates the mention can only be resolved by linking to a cell value or common sense reasoning.}
  \label{tab:spider-example}
\end{table*}

However, it is still difficult for current state-of-the-art models to fill in the skeletons with semantically correct entities, especially when they are required to generalize to unseen DB schemas~\cite{yu-etal-2018-spider,suhr-etal-2020-exploring}.
To predict the correct entity, the model should have a database (DB) schema grounded understanding of the NL question, which means that the model should be able to jointly learn the semantics in the NL question and the structured knowledge in a given database.
We formulate two types of entity generation problems, which can be addressed by the following two linking processes respectively.

\noindent\textbf{Schema linking.}
Schema linking~\cite{guo-etal-2019-towards,wang-etal-2020-rat} is an instance of entity linking~\cite{shen-2014-el} in the context of linking to relational DB schema. Text-to-SQL semantic parsers should learn to recognize an entity mention in the NL question and link it to the corresponding unique entity in the DB schema.
This task can be challenging due to the diversity and ambiguity NL mentions.
However, in practice, the solution is often relatively easy when a particular entity is well realized with similar wording in both the NL question and DB schema.
As shown in \autoref{tab:spider-example}, in Spider \cite{yu-etal-2018-spider}, the underlined mentions can almost exactly match the corresponding schema entities.
Therefore, current state-of-the-art parsers normally address this problem with simple string matching or embedding matching modules.

\noindent\textbf{Structural linking.} While some entities are generated because they are mentioned in the NL question, others can be generated because of their role as special functional components in SQL, \textit{e.g.}, \texttt{contid} and \texttt{stuid} in the \texttt{ON} clauses of the first and third examples in \autoref{tab:spider-example}.
These entities usually cannot find their corresponding mentions in the NL question but can be induced by the structural constraints of SQL.
Such phenomenons are generally referred to as the \textit{structural mismatch} between NL and formal languages~\cite{Kwiatkowski-etal-2013-scaling,Berant-liang-2014-semantic}.
This process frequently occurs when generating complex SQL queries. We propose to treat this entity generation process as finding a structural link between current candidates and past generated entities. Although previous work has considered some simple structural constraints~\cite{guo-etal-2019-towards}, to the best of our knowledge, we are the first to formally describe this process.

While schema linking and structural linking can complement each other (\textit{e.g.}, the entity used by the \texttt{GROUP BY} clause needs to be a column in a previously selected table and also has its mention in the NL question), they actually address different types of problems.
In most cases, \textit{e.g.}, the examples in \autoref{tab:spider-example} described before, a decoder may need to discriminate one from another for better generation performance.

In this work, we propose to use a \textit{dynamic gating} mechanism to serve as a switch between the two linking processes. 
Schema linking can be implemented as the decoder attending to the encoded representations of the NL question to find an entity mention, then locating the corresponding entity from the schema.
On the other hand, in structural linking, our decoder performs self-attention over those past decoder states where an entity was generated and further makes a decision between copying one of the previously generated entities or taking one of its linked entities (\textit{e.g.}, the foreign key of a previously joined table) based on the schema structure. From the model's perspective, it can also be viewed as a memory pointer network that enhances the structured prediction ability of an auto-regressive model, and the dynamic gating determines when to emphasize this enhancement. Our proposed method can be easily applied to most semantic parsers as long as their decoders explicitly or implicitly have two modules that deal with schema linking and structural linking.

We integrate the dynamic gating technique to two state-of-the-art Text-to-SQL parsing models~\cite{bogin-etal-2019-representing,bogin-etal-2019-global} and further augment them with pretrained BERT~\cite{devlin-etal-2019-bert} word representations.
We evaluate our model on the Spider dataset which is challenging because of its cross-domain setting where a model needs to generalize to not only complex SQL but also unseen DBs.
Experimental results show that our proposed method consistently yields an improvement of more than 3\% on exact set matching accuracy and sees the most benefits when generating complex SQL.
Further analysis confirms that the models are dynamically switching between the two linking processes.

\section{Approach}
\begin{figure*}
  \centering
  \includegraphics[width=\linewidth]{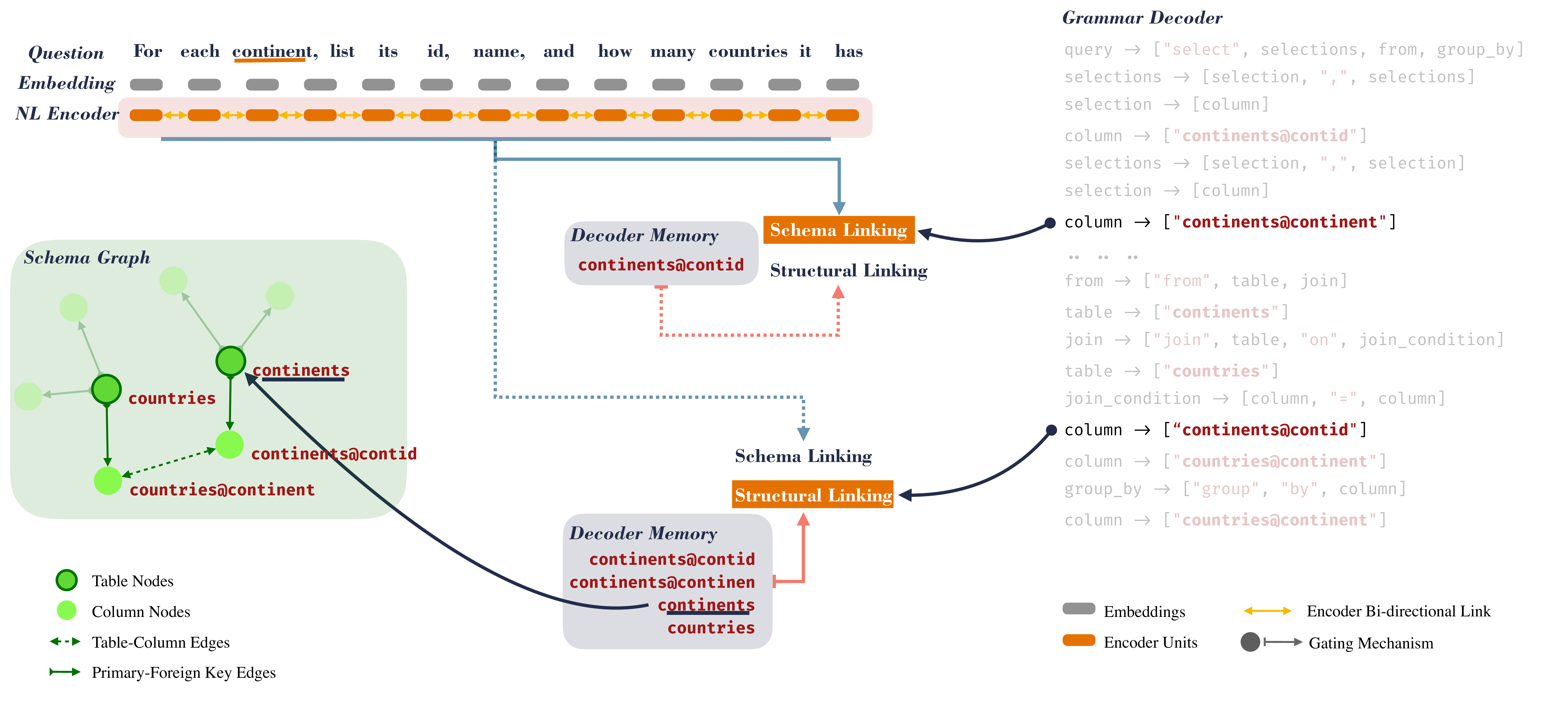}
  \caption{An illustration of our proposed method when running the first example shown in Table~\ref{tab:spider-example}. This figure shows two independent entity generation procedures, the top one favors schema linking while the bottom one favors structural linking. Some details are omitted for the sake of simplicity. Grammars are simplified to fit in the limited space, readers are encouraged to refer to~\protect\cite{krishnamurthy-etal-2017-neural} for details.}
  \label{fig:model}
\end{figure*}
In this section, we first formulate the Text-to-SQL semantic parsing task in \cref{sec:problem-definition}.
We will then describe the details of our proposed method in \cref{sec:dynamic-gating}.

\subsection{Text-to-SQL Semantic Parsing}
\label{sec:problem-definition}

The task of Text-to-SQL semantic parsing is to predict a SQL query $\mathcal{S}$ based on input $(\mathcal{Q}, \mathcal{G})$ where $\mathcal{Q}=\{q_1, \dots, q_{|\mathcal{Q}|}\}$ is the NL question and $\mathcal{G}=(\mathcal{V}, \mathcal{E})$ is the DB schema being queried.
In the schema, $\mathcal{V}=\{(e_1, t_1), \dots, (e_{|\mathcal{V}|}, t_{|\mathcal{V}|})\}$ is a set which usually contains two types of entities (\textit{i.e.}, tables and columns\footnote{Note that columns may have more fine-grained types like binary, numeric, string and date/time, primary/foreign etc.}) and their textual descriptions (\textit{i.e.}, table names and column names), while $\mathcal{E}=\{(e^{(s)}_1, e^{(t)}_1, l_1), \dots, (e^{(s)}_{|\mathcal{E}|}, e^{(t)}_{|\mathcal{E}|}, l_{|\mathcal{E}|})\}$ contains the relations $l$ between source entity $e^{(s)}$ and target entity $e^{(t)}$, \textit{e.g.}, table-column relationships, foreign-primary key relationships,\footnote{In SQL, a foreign key in one table is used to refer to a primary key in another table to link these two tables together for joint queries.} etc. The output $\mathcal{S}=\{a_1, \dots, a_{|\mathcal{S}|}\}$ is a sequence of decoder actions which further compose an AST of SQL.

\tikzset{
  treenode/.style = {shape=rectangle,
                     draw, align=center,
                     top color=white, bottom color=white},
  root/.style     = {treenode, font=\bfseries, text=white, top color=uva, bottom color=uva},
  env/.style      = {treenode, font=\normalsize},
  dummy/.style    = {circle,draw}
}

\begin{figure}
\centering
\begin{tikzpicture}
  [
    grow                    = right,
    sibling distance        = 4.5em,
    level distance          = 7em,
    edge from parent/.style = {draw, -latex},
    every node/.style       = {text width=3em, font=\footnotesize, midway, align=center},
    sloped
  ]
  \node [root] {Link gate}
    child { node [env] {NL\\encoder}
      edge from parent node [below] {Schema linking} }
    child { node [root] {Copy gate}
      child { node [env] {Link entity}
              edge from parent node [right, align=center] {}
              node [below] {}}
      child { node [env] {Copy entity}
              edge from parent node [right] {}
              node [above] {}}
              edge from parent node [above] {Structural linking} };
\end{tikzpicture}
\caption{An illustration of our gating mechanism.}
\end{figure}
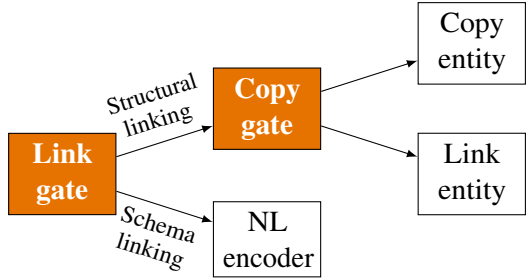

Typical state-of-the-art Text-to-SQL parsers, consist of three components: a NL encoder, a schema encoder and a grammar decoder~\cite{guo-etal-2019-towards,bogin-etal-2019-representing}.


The \textbf{NL encoder} takes the NL question tokens $\mathcal{Q}$ as input, maps them to word embeddings $\bm{E}_{\mathcal{Q}}$, then feeds them to a Bi-LSTM~\cite{hochreiter1997long}. The hidden states of the Bi-LSTM serve as the contextual word representation of each token.

The \textbf{schema encoder} takes $\mathcal{G}$ as input and builds a relation-aware entity representation for every entity in the schema.
The initial representation of an entity is a combination of its words embeddings and type information.
Then self-attention~\cite{zhang-etal-2019-editing,shaw-etal-2019-generating} or graph-based models~\cite{bogin-etal-2019-representing,wang-etal-2020-rat} are utilized to exploit the relational information between each pair of entities from the DB schema, thus produce the final representation of all entities $\bm{H}_{\mathcal{V}} \in \mathbb{R}^{|\mathcal{V}|\times\operatorname{dim}}$.
We will detail this in \cref{sec:gnn}.

Finally, a \textbf{grammar decoder}~\cite{xiao-etal-2016-sequence,yin-neubig-2017-syntactic,krishnamurthy-etal-2017-neural} generates an AST of output SQL in a depth-first order.
The decoder is typically an auto-regressive model (\textit{e.g.}, LSTM) which estimates the probability of generating an action sequence.


There are \emph{two cases} of using actions. 
Depending a specific case, 
an action is either
\begin{enumerate*} [label=(\roman*)]
\item \label{itm:rule} producing a new production rule to unfold the leftmost non-terminal node in the AST, or
\item \label{itm:ent} generating an entity (\textit{e.g.}, a table or a column) from the DB schema if it is required by last output production rule.
\end{enumerate*}
In the former case, at step $t$, the decoder normally uses its hidden states $\bm{h}_t$ to retrieve a context vector $\bm{c}_t$ from the NL encoder. Then an action embedding $\bm{a}_t$ is produced based on the concatenation of $\bm{h}_t$ and $\bm{c}_t$. This action embedding will directly predict a production rule from the target vocabulary which is a subset of a fixed number of production rules.
For the latter case, the decoder needs to estimate a probability distribution over a schema-specific vocabulary under grammatical constraints which come from the structure of both the output SQL and the DB schema, as well as semantic constraints implied in the NL question.

\subsection{Dynamic Gating}\label{sec:dynamic-gating}

In this paper, we focus on the decision made when the decoder is looking for an entity to fill in a slot (\textit{i.e.}, case \ref{itm:ent} in the last paragraph). Our decoder predicts the entity based on a mixed probability model consisting of two processes:
\begin{itemize}
    \item \textbf{Schema linking.} The decoder attends to the output of the NL encoder (which can be seen as selecting a most relevant NL mention), then finds the corresponding entity based on string-matching or embedding-matching results.
    \item \textbf{Structural linking.} The decoder self-attends to the output states from those previous decoding steps which have generated entities, then finds another entity which is structurally linked to the attended entity. 
\end{itemize}
The choice between them is controlled by a gating mechanism called the \textit{link gate}.

Formally, the marginal probability of generating an entity $e$ is defined as follows:
\begin{equation}
\begin{split}
    \operatorname{Pr}(a_t = e) &=
    \operatorname{Pr}(e|\textsc{Schm})\operatorname{Pr}(\textsc{Schm})
    + \operatorname{Pr}(e|\textsc{Strct})\operatorname{Pr}(\textsc{Strct})
\end{split}
\end{equation}
where $\operatorname{Pr}(\textsc{Schm})$ and $\operatorname{Pr}(\textsc{Strct})$ are the probability of choosing schema linking and structural linking respectively. They are further computed as:
\begin{align}
\label{eqn:gating}
    \rho_{\operatorname{link}} &= \operatorname{Sigmoid}(\operatorname{FF}(\bm{a}_t)) \\
    \operatorname{Pr}(\textsc{Schm}) &= \rho_{\operatorname{link}} \\
    \operatorname{Pr}(\textsc{Strct}) &= 1 - \rho_{\operatorname{link}}
\end{align}
\autoref{eqn:gating} stands for our proposed link gate which is computed by the action embedding $\bm{a}_t$. The reason for purely basing the gate value on $\bm{a}_t$ is that intuitively the choice between the two processes is about the role of the current entity we want to generate. The role of an entity is determined by the SQL clause that contains it. Since $\bm{a}_t$ is directly used to predict a production rule in case~\ref{itm:rule}, it should be able to capture this information. The link gate allows the decoder to dynamically choose between information from our two linking processes, and prevents them from interfering each other.

In practice, we model the probability of the schema linking process generating an entity mentioned in the NL question, namely $\operatorname{Pr}(e|\textsc{Schm})$, as a multiplication of the attention weights $\bm{\lambda} \in \mathbb{R}^{|\mathcal{Q}|}$ over the NL encoder outputs and a schema linking matrix $M\in \mathbb{R}^{|\mathcal{Q}|\times|\mathcal{V}|}$. 
The probability of the structural linking process, namely $\operatorname{Pr}(e|\textsc{Strct})$, is similarly computed by multiplying decoder self-attention weights and a structural linking matrix $T\in \mathbb{R}^{|\mathcal{V}|\times|\mathcal{V}|}$.

The structural linking matrix $T$ captures the relationship between every pair of entities given the relational DB schema. Common structural links include relations between a table and its columns, a table and its primary/foreign key, a primary key and one of its linked foreign keys in other tables, etc. There are also multi-steps links which are the combinations of the one-step links listed above.
Note that there may not be a unique link between every pair of entities and some entities may not have a link between them at all.
Meanwhile, an entity can link to itself which can be considered to be a special zero-step structural link. It is the only structural link modeled in most current Text-to-SQL semantic parsers.

We compute the structural linking score between an entity $e_i$ and $e_j$ by an additive attention mechanism~\cite{bahdanau2014neural}, as follows:
\begin{equation}
T_{i, j}=\mathbf{v}_{\alpha}^{\top} \tanh \left(\mathbf{W}_{\alpha}\left[\bm{e}_{i} ; \bm{e}_{j}\right]\right)
\label{eqn:struct-matrix}    
\end{equation}
where $\bm{e}_i$ and $\bm{e}_j$ are the corresponding entity representations retrieved from $\bm{H}_\mathcal{V}$, while $\mathbf{v}_{\alpha}$ and $\mathbf{W}_{\alpha}$ are both trainable parameters. Compared to the dot-product attention used in \newcite{bogin-etal-2019-representing} and \newcite{bogin-etal-2019-global}, the additive attention we used here is expected to capture more of the structural relationship between entity pairs rather than only the similarity of entity representations.

$T$ is expected to capture all types of relationships between entities, but it can be overwhelmed by the large workload. So, we single out the zero-step relationship (\textit{i.e.}, copying) and address it by another structural linking matrix $T_{\operatorname{copy}}=\bm{I}$, which is trivially an identity matrix in this case. To choose between copying and other types of links, a \textit{copy gate} ($\rho_{\operatorname{copy}}$) is obtained in the same manner as to how we compute the link gate in~\autoref{eqn:gating}.

We use decoder self-attention to find the past generated entity which could have structural constraints on the entity that we currently want to generate.
\begin{gather}
\bm{\beta} = \operatorname{Softmax}(\operatorname{Attention}(\bm{a}_t, \bm{H}_{m})) \\
\bm{H}_{m} = \{\bm{a}_i|i<t, a_i\in e\} \label{eqn:memory}
\end{gather}
$\bm{H}_{m}$ is a memory matrix consisting of the action embeddings from every past decoding step which has generated an entity. 
In this way, we compute two sets of attention weights $\bm{\beta}_{\operatorname{copy}}$ and $\bm{\beta}_{\operatorname{link}}$ for copying and linking separately using the additive attention~\cite{bahdanau2014neural} again. The motivation for separate attention weights is that these two linking patterns might need to attend to different generated entities.

Overall the probability of generating an entity via structural linking is modeled as:
\begin{equation}
    \begin{split}
\operatorname{Pr}(a_t=e_i|\textsc{Strct}) = \rho_{\operatorname{copy}}(\bm{\beta}_{\operatorname{copy}}T_{\operatorname{copy}})_{i}
+(1-\rho_{\operatorname{copy}})(\bm{\beta}_{\operatorname{link}}T)_i
    \end{split}
\end{equation}
Finally, this probability is mixed with the probability of schema linking controlled by the link gate.


\section{Model Implementation}\label{sec:gnn}

In this section, we describe how we integrate our proposed method into a grammar decoder and leverage the entity representation from a GNN module.

We use the type constrained grammar decoder from \newcite{krishnamurthy-etal-2017-neural}. To predict $a_t$ at time step $t$, the decoder will first obtain the context vector $\bm{c}_t$ from the NL encoder by performing dot-product attention~\cite{luong-etal-2015-effective}.
Then the action embedding is generated by a feed-forward network taking the concatenation of decoder hidden state and context vector as input.
\begin{equation}
\bm{a}_t = \operatorname{FF}([\bm{h}_t; \bm{c}_t])
\label{eqn:action}
\end{equation}
$\bm{a}_t$ is used to predict the production rule or estimate the gate values in the entity generation process.

We adopt the idea from \cite{bogin-etal-2019-representing,bogin-etal-2019-global} to learn a schema relation-aware entity representation $\bm{H}_{\mathcal{V}}$ by a GNN module.\footnote{We choose these models because of the ability of GNNs to model various types of structural links, and they are among a few state-of-the-art models that are publicly available at the time of writing.} The initial embedding of each entity $h_e^{(0)}$ is defined as a non-linear transformation of the combination of its type embedding and the average over the word embeddings of its neighbors in the schema graph.
In later time steps, the hidden state is updated by a gated recurrent unit~\cite{cho-etal-2014-learning,li2015gated} as $h_{e}^{(l)}=\operatorname{GRU}\left(h_{e}^{(l-1)}, x_{e}^{(l)}\right)$, where the input $x_{e}^{(l)}$ is defined as a weighted summation over the hidden states of its neighbor entities:
$$x_{e}^{(l)}=\sum_{t\in\{\leftarrow, \rightarrow,\leftrightarrow\}} \sum_{(s, e, l) \in \mathcal{E}, l=t} \mathbf{W}_{t} h_{s}^{(l-1)}+b_{t}$$
They consider three edge types, \textit{i.e.}, bidirectional edges between a table and its contained columns $\leftrightarrow$, unidirectional edges between a foreign key and a connected primary key $\leftarrow$ and its reverse version  $\rightarrow$.
Given a fixed GNN recurrence step $L$, we have the final hidden states of all the entities in the graph as the entity representation $\bm{H}_\mathcal{V}=\{h_{e}^{(L)}|(e, t)\in \mathcal{V}\}$.
We also adopt their schema linking module to create a schema linking matrix $M$ based on word embedding similarity and some simple manually design features (\textit{e.g.}, editing distance and lemma).
In their \textsc{GlobalGNN}~\cite{bogin-etal-2019-global}, an additional GNN and an auxiliary training loss are added to filter out irrelevant nodes in the graph, thus producing a better entity representation.

To augment our model with pretrained BERT embeddings, we follow \newcite{hwang2019comprehensive} and \newcite{zhang-etal-2019-editing} to feed the concatenation of NL question and the textual descriptions of DB entities to BERT and use the top layer hidden states of BERT as the input embeddings.
\section{Experiments}

We evaluate the effectiveness of our proposed method by integrating it into two state-of-the-art semantic parsers on the Spider dataset and further ablate out some components to understand their contributions.

\subsection{Experiment Setup}

We implement our model using PyTorch~\cite{NEURIPS2019_9015} and AllenNLP~\cite{gardner-etal-2018-allennlp}. For the GNN and \textsc{GlobalGNN} models we revise and build upon the code released in \cite{bogin-etal-2019-representing,bogin-etal-2019-global}.
We re-ran the experiment and report the results on our re-implementation and found our results slightly improves upon their reported results. In BERT experiments, we use the base uncased BERT model with 768 hidden size provided by HuggingFace's Transformers library~\cite{Wolf2019HuggingFacesTS}. We follow the database split setting of Spider, where any databases that appear at testing time are ensured to be unseen at training time.
Our code and models are available at \url{https://github.com/sanxing-chen/linking-tale}.


    
\begin{table}[tb!]
\minipage{0.3\linewidth}
    \begin{tabular}{lr}
      \toprule
      Hardness & \# Example \\ \midrule
      Easy & 250 \\
      Medium & 440 \\
      Hard & 174 \\
      Extra & 170 \\
      All & 1034 \\
      \bottomrule
    \end{tabular}
  \caption{Number of examples in the development set of Spider with different hardness levels associated with the SQL need to be generated.}
  \label{tab:spider}
\endminipage\hfill
\minipage{0.66\linewidth}
  \begin{tabular}{lccccc}
      \toprule
      Model &  Acc. & Easy & Medium & Hard & Extra \\ \midrule
      GNN & 47.7\% & 68.8\% & 51.8\% & 31.2\% & 22.9\% \\
      \hspace{3mm} + Ours & 50.7\% & 66.4\% & 54.8\% & 42.8\% & 25.3\% \\
      \hline
      \textsc{GlobalGNN} & 49.3\% & 69.2\% & 53.0\% & 32.8\% & 27.6\% \\
      \hspace{3mm} + Ours & 52.8\% & 70.4\% & 55.7\% & 46.6\% & 25.9\% \\
      \hspace{3mm} + BERT & 53.5\% & 76.0\% & 57.3\% & 36.2\% & 28.3\% \\
      \hspace{3mm} + BERT + Ours & 57.6\% & 73.6\% & 61.6\% & 48.9\% & 32.9\% \\
      \bottomrule
    \end{tabular}
  \caption{Exact Set Matching Accuracy on SQL queries with different hardness levels in the development set of Spider.  Greatest improvements in the Hard level; small fluctuation in Easy level due to gate bias.}
  \label{tab:exp}
\endminipage
\end{table}

\subsection{Experimental Results} 


The experimental results in \autoref{tab:exp} show that our proposed gating mechanism leads a substantial improvement on all the GNN, \textsc{GlobalGNN}, and BERT baselines. Spider questions are divided into different levels of difficulty (hardness). Most of the improvements come from gains in complicated (\textit{i.e.}, Medium, Hard and Extra Hard) SQL generation. Specially, we observe up to $13.8\%$ gains in the Hard set when applying our method on the \textsc{GlobalGNN} baseline. One major contribution comes from the partial matching F1 score of \texttt{IUEN} (\textit{i.e.}, SQL clauses \texttt{INTERSECT}, \texttt{UNION}, \texttt{EXCEPT}, \texttt{NESTED} which only appear in Hard and Extra Hard levels) increasing from $25.4\%$ to $39.7\%$. We also notice that the SQL output well-formedness is improved. For instance, before applying our method the decoder would occasionally select the same columns twice to perform the \texttt{ON} clause.\footnote{This is different from the case of intentionally self join.} After applying our dynamic gating, this issue is virtually eliminated (error rate from $2\%$ to $0.2\%$).\footnote{Although this issue can be resolved by engineering more grammar rules, we leave it as an indicator of the improvement of the well-formedness of the output SQL.}

\begin{wrapfigure}{r}{0.5\textwidth}
    \centering
  \includegraphics[width=0.48\textwidth]{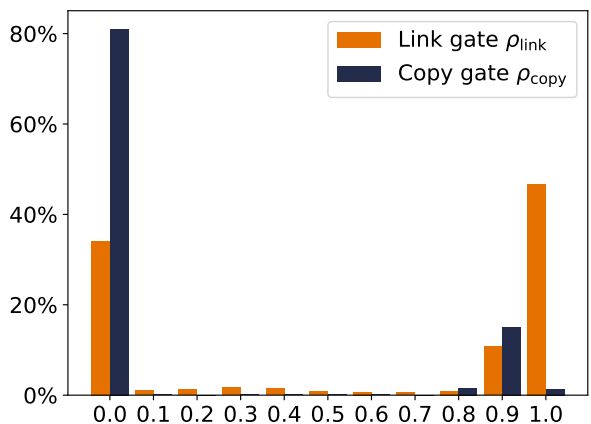}
  \caption{Value distributions of the link gate and copy gate measured in dev set of Spider using the \textsc{GlobalGNN} model. Values of copy gate are considered when the corresponding link gate value is small ($\rho_{\operatorname{link}}<0.1$).}
  \label{fig:gate_value}
\end{wrapfigure}

As shown in \autoref{fig:gate_value}, the values of both gates are polarized to $0$ or $1$, thus making the gating mechanism act as a binary gate. These statistics coincide with our hypothesis that \emph{most entity generation decisions in Text-to-SQL can be solely made by evidence from either schema linking or structural linking}. In addition, among all the cases where structural linking is chosen and the copy gate takes control, fewer than 20\% of cases favor copying. This suggests that there are lots of circumstances where different kinds of structural linking are adopted.

\subsection{Alternative Approaches and Ablation}

We also conduct several experiments to examine several design choices in our proposed method.



\noindent\textbf{Sharing Action Embedding.} In the design of our gating mechanism, one critical decision is to use the action embedding $\bm{a}_t$ to perform decoder self-attention and produce the gating values. This is based on our intuition that the action embedding captures the structural information of the output SQL at the current position.
To verify this decision, we conduct an ablation experiment by using a \textit{dedicated embedding} to produce the gating value. This dedicated embedding is produced in exactly same way as we generated $\bm{a}_t$ in \autoref{eqn:action}, but uses a different set of parameters for the feed-forward network. As we can see from \autoref{tab:ablation} (``dedicated embed"), sharing the parameters with the action embedding is important.

\noindent\textbf{Keeping entities.} 
\newcite{guo-etal-2019-towards} also uses a memory-augmented pointer network to perform a copy mechanism which assists column selection. In contrast with our memory matrix consisting of action embeddings (\autoref{eqn:memory}), their memory matrix consists of the entity embeddings of columns that have been selected previously. They further remove the columns from the candidates in the schema linking process once they are generated to prevent the decoder from repeatedly generating the same columns. To determine if our dynamic gating and structural linking module can add enough structural constraints to the decoding process to resolve this kind of problem, we conduct an experiment where we also remove the entities in the schema linking process after they are generated (\textit{i.e.}, ``removing entity" in \autoref{tab:ablation}) to see if it further improves the model. Our results shows that this change can actually hurts the model. Specifically, we observe a drop in accuracy of the \texttt{WHERE} and \texttt{IUEN} clauses, which suggests that in our context, the information about a specific entity in the schema linking process is still useful even after the entity has been generated once.

\begin{wrapfigure}{r}{0.55\textwidth}
\centering
\setlength{\tabcolsep}{3pt}
\begin{tabular}{lrr}
    \toprule
    \multirow{2}{*}{Model} &  \multicolumn{2}{c}{Dev Acc. (\%)} \\ 
    & \textsc{GlobalGNN} & \textsc{GlobalBERT}  \\\midrule
    Base & 49.3 & 53.5 \\ \hline
    Ours & \textbf{52.8} & \textbf{57.6} \\
    \hspace{3mm}dedicated embed & 50.0 & 55.4 \\
    \hspace{3mm}removing entity & 49.4 & 57.1 \\
    \hspace{3mm}without copy & 50.9 & 55.8 \\
    \bottomrule
  \end{tabular}
  \captionof{table}{Alternative approaches and ablation results.}
  \label{tab:ablation}
\end{wrapfigure}

\noindent\textbf{Copy Gate.} In addition, removing the copy gate and copy mechanism also harms the performance of our model (\textit{i.e.}, ``without copy" in \autoref{tab:ablation}). This result confirms that it is beneficial to handle different types of structural links separately. We hypothesize that different types of structural links conflict with each other, so they are hard to fit in one structural linking matrix.
Overall, these two results further supports our claim that the copy gate can determine when to copy or link to an entity by itself.

\begin{table*}[tb!]
  \centering
  \begin{tabular}{p{0.05\linewidth}p{0.85\linewidth}}
    \toprule
    NL & \textit{Show the stadium name and the number of concerts in each stadium.} \\ \hline
    SQL & \begin{gatelstlisting}
 select@*\txtunderset{\flatgcv{\text{N/A}}{\text{N/A}}}{stadium.name}*@, count(*) from @*\txtunderset{\flatgcv{0.15}{0.00}}{concert}*@ join @*\txtunderset{\flatgcv{0.00}{0.00}}{stadium}*@ 
 on @*\txtunderset{\flatgcv{0.09}{0.00}}{concert.stadium\_id}*@ = @*\txtunderset{\flatgcv{0.00}{0.00}}{stadium.stadium\_id}*@ group by @*\txtunderset{\flatgcv{0.00}{0.96}}{concert.stadium\_id}*@;\end{gatelstlisting}\\ 
    \midrule
    NL & \textit{Which city has most number of departing flights?} \\ \hline
    SQL & \begin{gatelstlisting}
 select@*\txtunderset{\flatgcv{\text{N/A}}{\text{N/A}}}{airports.city}*@ from @*\txtunderset{\flatgcv{0.82}{0.01}}{airports}*@ join @*\txtunderset{\flatgcv{0.00}{0.00}}{flights}*@ on  @*\txtunderset{\flatgcv{0.00}{0.00}}{flights.airportcode}*@ = @*\txtunderset{\flatgcv{0.00}{0.00}}{flights.sourceairport}*@ group by @*\txtunderset{\flatgcv{0.50}{1.00}}{airports.city}*@ order by count (*) desc limit 1\end{gatelstlisting}\\ 
    \bottomrule
  \end{tabular}
  \caption{Sample predictions of our model. $\rho_{\operatorname{link}}$ and $\rho_{\operatorname{copy}}$ are abbreviated as $\rho_{\operatorname{l}}$ and $\rho_{\operatorname{c}}$ respectively. Gate values are not applicable (denoted by N/A) for the first entity since it has no previously generated entity. Some of the results reflect gate bias, see text for details.}
  \label{tab:sample}
\end{table*}

\subsection{Error Analysis and Discussion}


\noindent\textbf{Gate bias.} Error analysis reveals that our gating mechanism is biased, \textit{e.g.}, for the first few entities being selected in a SQL query the link gate is trained to favor schema linking in most cases. But, such bias could sometimes be wrong. In such cases where structural linking is needed but absent, the model may select duplicate columns or the wrong table during decoding.
Similarly, the copy gate might be biased toward copying an entity from memory in \texttt{GROUP BY} clauses. Out of all the SQL clause components, only the \texttt{GROUP BY} clause's partial matching F1 score drops (about $3\%$) due to this copy gate bias.
It is true that the entity needed in the \texttt{GROUP BY} clause is usually selected, but the information from schema linking can still be beneficial,\footnote{The semantics of some pronouns (\textit{e.g.}, ``each" and ``which" in the examples of \autoref{tab:sample}) in NL question match with the \texttt{GROUP BY} clause, but this could be a dataset bias.} \textit{e.g.}, in the second example of \autoref{tab:sample}, the model wants to copy the wrong entity but the link gate rectifies it with schema linking information. Our gating mechanism only relies on the current action embedding (which can be seen as short-term structural information) to determine the gate values. We believe that introducing more global structural constraints is a promising direction to find a more flexible and accurate gating mechanism.

\noindent\textbf{Short attention spans.} In our experiments, we notice that the action history the model usually attends to is very short, \textit{i.e.}, the model only utilizes the the output memory of the most recent three entities in 99\% of cases. This coincides with similar findings in language modeling~\cite{daniluk2017frustratingly} where the augmented-memory was expected to facilitate the modeling of long-range dependencies but failed to do so. Although long-term context is important for language modeling, it is not as important in our Text-to-SQL scenario since most dependencies in programming languages like SQL lie within a short span. We are interested in exploring semantic parsing tasks which requires long-term structural constraints using our method in the future.

\noindent\textbf{Structural linking patterns.} We have shown the effectiveness of our method in dealing with different types of structural links separately using different components of the model in the previous section. So far, the only special type of structural links we can explicitly model is the copy mechanism, and we treat all other types of links uniformly using additive attention in \autoref{eqn:struct-matrix}. This might limit the model's ability to take advantage of the complicated relationships between entities. Currently, the entity representation provided by the GNN model is still difficult to explain, because the node representations contain a mix of information from different message-passing steps. One could imagine training GNNs with different message-passing steps each modeling a different level of structural linking, could lead to a more clear and expressive linking pattern.
\section{Related Work}
\noindent{\bf Semantic parsing.} Semantic parsing research focus on mapping NL to formal languages like lambda calculus ~\cite{zettlemoyer-2005-learning,kwiatkowksi-etal-2010-inducing,liang-etal-2011-learning,dong-lapata-2016-Language}, Prolog-style queries~\cite{Zelle-1996-Learning,tang-mooney-2000-automated}, and more recently to SQL~\cite{warren-pereira-1982-efficient,popescu2003towards,giordani2009semantic,zhongSeq2SQL2017,iyer-etal-2017-learning}.  It can also tackle the problem of parsing NL descriptions to complicated general-purpose programming language such as Python~\cite{Ling-etal-2016-Latent,Rabinovich-etal-2017-Abstract,yin-neubig-2017-syntactic}. Our proposed method is tested for Text-to-SQL parsing and can be adapted to other semantic parsing applications.

\noindent{\bf Structural mismatch.} Programming languages like SQL express the same intent in a completely different way from NL by design~\cite{kate-2008-transforming}. The phenomenon called structural mismatch widely exists between NL and various programming language and is a major challenge in semantic parsing~\cite{dong_learning_2019}.
To alleviate the structural mismatch problem, 
early approaches rely on linguistic formalisms like parsing results from flexible CCGs~\cite{zettlemoyer-2005-learning,zettlemoyer-collins-2007-online,Kwiatkowski-etal-2011-lexical,Kwiatkowski-etal-2013-scaling}.
\newcite{chen-etal-2016-sentence} proposed to use sentence rewriting to revise the NL question to a new question which has the same structure with the targeted logical form.
Recently, \newcite{guo-etal-2019-towards} proposed to first translate the NL question to an intermediate representation (IR) designed to bridge NL and SQL, then use a deterministic algorithm to convert the IR to SQL. In addition to taking a considerable amount of engineering effort, their designed IR is still unable to cover some SQL grammars like the self-join in the \texttt{ON} clause, and is more challenging to apply to other programming languages. We deal with this problem by explicitly modeling the prediction structure with external predefined structure (\textit{i.e.}, DB schema) by structural linking.

\noindent{\bf Memory pointer network.} Memory networks were first introduced in the context of the question answering task, where they served as a differentiable long-term knowledge base to enhance an auto-regressive model's poor memory~\cite{weston2014memory,sukhbaatar2015end}.
Copy mechanisms use attention as a pointer to select and copy items from source text, thus addressing the problem of a variable output vocabulary size~\cite{vinyals2015pointer,see-etal-2017-get}.
Recent research has applied memory-augmented pointer networks to various NLP tasks, including task-oriented dialogue~\cite{wu2019global} and also semantic parsing~\cite{liang-etal-2017-neural,guo-etal-2019-towards}. Our dynamic gating mechanism can also be seen a memory controller, except our memory is read-only and acts as both the query and key in a pointer network. Different from most current techniques, our pointer network does not only perform copying but can also point to a start point of structural linking.
\section{Conclusion}
In this paper, we formulated the entity generation process in Text-to-SQL semantic parsing as two kinds of linking problems, namely schema linking and structural linking. We further proposed a dynamic gating mechanism to explicitly model the decision between these two linking processes. Experimental results show the effectiveness of our proposed method and confirm our intuitions. In the future, we would like to apply our proposed method to other semantic parsing tasks, such as general purpose code generation, where structural constraints may be more important.

\section*{Acknowledgments}
We thank Yu Bai and members of the UVA NLP group for valuable discussion and feedback. We also thank all anonymous reviewers for their helpful comments and suggestions.


\bibliographystyle{coling}
\bibliography{anthology, coling2020}

\end{document}